\pdfoutput=1

\documentclass[11pt]{article}

\usepackage[]{acl}

\usepackage{times}
\usepackage{latexsym}
\usepackage{hyperref}
\usepackage{microtype}
\usepackage{amsmath,amssymb}
\usepackage{epsfig,graphicx,subfigure,caption}
\usepackage{algpseudocode}
\usepackage[normalem]{ulem}
\usepackage{amsmath}
\usepackage[linesnumbered,algoruled,boxed,noend]{algorithm2e}
\usepackage{xcolor}
\usepackage{color, colortbl}
\usepackage{multirow,booktabs, hhline}
\usepackage{amsmath, bm}
\usepackage[linesnumbered,algoruled,boxed,noend]{algorithm2e}
\usepackage{wrapfig}
\usepackage{adjustbox}
\usepackage{makecell}
\usepackage{tabularx}
\usepackage{enumerate}
\usepackage{lipsum}
\usepackage{graphicx}
\usepackage[shortlabels]{enumitem}

\definecolor{darkgreen}{RGB}{0,100,0}

\usepackage[T1]{fontenc}

\usepackage[utf8]{inputenc}

\usepackage{microtype}

\title{Can Language Model Moderators Improve the Health of Online Discourse?
\\ \small{\textcolor{red}{\textit{Warning: this paper contains and discusses content that may be offensive.}}}
}

\author{
Hyundong Cho\thanks{\xspace~Corresponding author.}, Shuai Liu, Taiwei Shi, Darpan Jain, Basem Rizk, Yuyang Huang, \\ \textbf{Zixun Lu, Nuan Wen, Jonathan Gratch, Emilio Ferrara}, and  \textbf{Jonathan May}\\
Department of Computer Science and Information Sciences Institute \\
University of Southern California \\
\small{\texttt{hd.justincho@gmail.com}}
}

\begin{document}
\maketitle

\renewcommand{\arraystretch}{1.5}

\begin{abstract}

Conversational moderation of online communities is crucial to maintaining civility for a constructive environment, but it is challenging to scale and harmful to moderators. The inclusion of sophisticated natural language generation modules as a force multiplier to aid human moderators is a tantalizing prospect, but adequate evaluation approaches have so far been elusive. In this paper, we establish a systematic definition of conversational moderation effectiveness grounded on moderation literature and establish design criteria for conducting realistic yet safe evaluation. We then propose a comprehensive evaluation framework to assess models' moderation capabilities independently of human intervention. With our framework, we conduct the first known study of language models as conversational moderators, finding that appropriately prompted models that incorporate insights from social science can provide specific and fair feedback on toxic behavior but struggle to influence users to increase their levels of respect and cooperation.

\end{abstract}
\section{Introduction}

\begin{figure}[t]
    \centering
    \includegraphics[width=\linewidth]{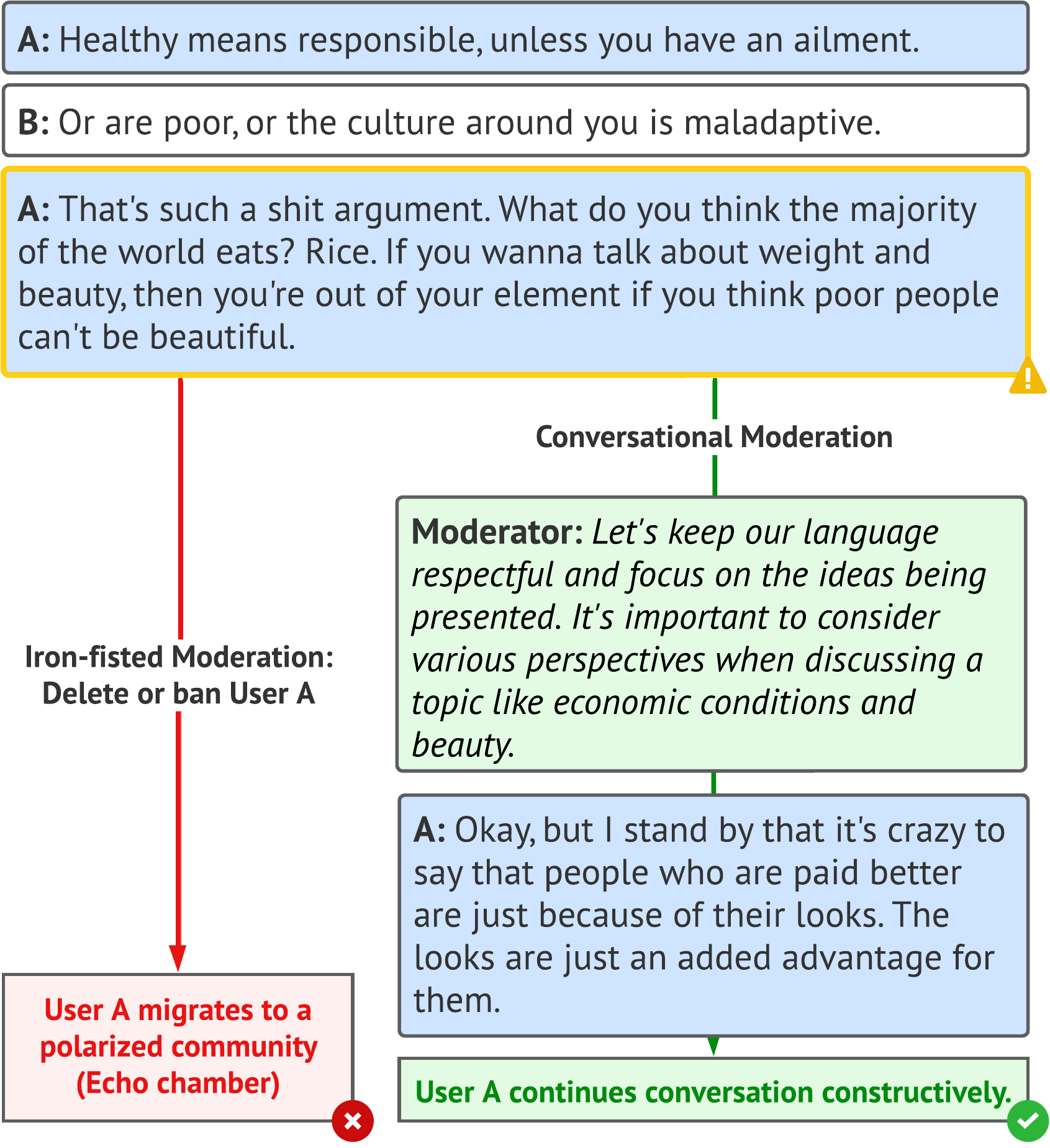}
    \caption{While banning users or deleting their comments may push them towards echo chambers (\textcolor{red}{left}), conversational moderation can guide users towards more constructive behavior (\textcolor{darkgreen}{right}). 
    Recent developments in instruction-tuned language models with conversational capabilities present an opportunity to perform conversational moderation at scale and improve the health of online discourse.}
    \label{fig:conversational_moderation_example}
\end{figure}

The rapid increase in online users and the growing polarization of society have created significant challenges in maintaining civil discourse and mitigating harmful content in online platforms~\cite{guberman2016quantifying, almerekhi2020investigating}. 
Effective moderation is necessary to counter this trend, but scaling moderation efforts to meet the demands of an expanding user base is difficult without some form of automation. 
Previous automatic moderation efforts have largely focused on banning or deleting comments from harmful users~\cite{srinivasan2019content, park-etal-2021-detecting-community}. 
However, such iron-fisted approaches can inadvertently push these users towards echo chambers that exacerbate polarization~\cite{cinelli2020echo}.

An alternative to these efforts is \textit{``conversational moderation,''} in which a moderator converses with the problematic user to guide discussions towards a more constructive outcome, as shown in \autoref{fig:conversational_moderation_example}. 
Recent studies have shown that engaging in conversations is an effective approach for moderating users' behaviors~\cite{combs2022anonymous}, and there are efforts such as \textit{The Commons}\footnote{\url{https://howtobuildup.org/programs/digital-conflict/the-commons-project/}} that encourage human moderators to interactively intervene 
 in controversial conversations on race and politics.  
However, human moderators have reported steep learning curves in engaging with harmful users~\cite{laurenson2019polarisation} and moderation is mentally taxing \cite{10.1145/3411764.3445092}, making this approach challenging to scale.
Therefore, human moderators could benefit from a reduced cognitive load by the availability of conversational moderation suggestions provided by \textit{moderator bots}. 
Fortunately, the recent advancement in prosocial dialogue models~\cite{kim-etal-2022-prosocialdialog, kim-etal-2023-soda} and instruction-tuned language models (ITLM) that can generalize well to new tasks with zero or very little task-specific data presents a potential for scaling up conversational moderation with the help of these suggestions~\cite{wei2021finetuned, kojima2022large, ouyang2022training,  bai2022training, kopf2023openassistant, zhou2023lima}. 
The central question is whether there is enough evidence that the moderator bot suggestions are effective enough for human moderators to use in the course of their duties. 

To this end, we seek to answer the following research questions in this paper:
\vspace{-0.15cm}

\begin{enumerate}[label={\bfseries R\arabic*:}, , itemsep=0pt]

\item How should we define and evaluate \textit{effectiveness} in conversational moderation?
\item How effective are moderator bots at conversational moderation?
\end{enumerate}
\vspace{-0.15cm}

To answer these questions, we develop an evaluation protocol driven from existing literature \cite{grimmelmann2015virtues, srinivasan2019content, kiesler2012regulating, argyle2023leveraging} and conduct  studies to determine the effects of conversational moderation on social cohesion and conflict resolution. 
We build a novel framework that enables us to monitor realistic user interaction with moderators in the context of a real, controversial online discussion, yet in a manner that ensures minimal risk to users, and
measure the perceived effect of moderation on user behavior via surveys given to users after a moderation encounter. 
With this framework, we evaluate a range of approaches to moderation, including existing prosocial dialogue models~\cite{kim-etal-2022-prosocialdialog, kim-etal-2023-soda} and prompted ITLMs informed by conflict resolution~\cite{lytle1999strategic}, cognitive behavioral therapy~\cite{clark2015socratic}, and prosocial communication techniques~\cite{rosenberg2015nonviolent}.

Our results show that prompted ITLM moderators can provide specific and fair feedback, but making users more respectful and cooperative is challenging. 
They largely outperform prosocial dialogue models, and one that incorporates Socratic dialogue techniques from cognitive behavioral therapy performs the best among the methods we explored. 
Interestingly, we also find that the perceived effectiveness of our moderators varies based on whether the evaluator is the moderated user or an observer.
To encourage the research community to build on our evaluation framework and study, we release our dataset of controversial conversations and completed conversations with annotations.\footnote{Available at \url{https://github.com/isi-nlp/isi_darma}}

\section{Evaluating Conversational Moderation}


Prior to exploring how to develop conversational agents that can assist with conversational moderation, we first define moderation effectiveness and its metrics in the context of conversational moderation (Section \ref{sec:definition}-\ref{sec:metrics}) and identify experimental design criteria that should be balanced (Section \ref{sec:design}. 
Then, based on these factors, we propose a framework for evaluating how useful language models (LM) can be in assisting with conversational moderation (Section \ref{sec:conversational_stubs}-\ref{sec:survey_questions}).

\begin{table*}[h]
\begin{adjustbox}{max width=\linewidth}
\renewcommand{\arraystretch}{1.5}
    \centering
    \begin{tabular}{p{0.1\linewidth}p{0.3\linewidth}p{0.8\linewidth}}
        \toprule 
        Criteria & Description & Implementation \\ \midrule
        Safe & Minimize harms. & Evaluation takes place offline with participants that provided consent and does not affect real users. \\ 
        Realistic & Replicate realistic context. & Start with controversial conversation stubs to replicate highly charged conversations. \\ 
        Controlled & Isolate and measure effect of moderation strategy. & Evaluate moderator bots with dyadic conversations between human participant and the bot, followed by a survey. \\ 
        \bottomrule
    \end{tabular}
\end{adjustbox}
    \caption{Experimental design criteria and their corresponding implementation that grounds our conversational moderation evaluation framework. We strike a balance in keeping our evaluation procedure safe, realistic, and controlled for evaluating the effectiveness of conversational moderation by LM moderators.}
    \label{tab:criteria_implementation_overview}
\end{table*}

\subsection{Definition of moderation effectiveness}
\label{sec:definition}

At the highest level, \citet{grimmelmann2015virtues} states that moderation should \textit{prevent abuse} and \textit{facilitate cooperation} for not only the moderated user, but also for other community members. 
In other words, moderation events should (i) make users more engaged and willing to cooperate (\texttt{cooperative}) and (ii) make them more respectful and less abusive (\texttt{respectful}). 

In addition, \citet{kiesler2012regulating} found that fairness and clarity of community rules decreases resistance to moderation procedures. 
Therefore, for a moderator to be effective, it must also (iii) be fair to all users involved in the conversation (\texttt{fair}) and (iv) make specific and relevant suggestions to facilitate cooperation (\texttt{specific}). 

\subsection{Metrics for conversational moderation effectiveness}
\label{sec:metrics}

\citet{srinivasan2019content} map the aforementioned moderation goals in Section \ref{sec:definition} to quantifiable metrics to measure a moderation event's effectiveness through its causal effect, such as the rate of noncompliance, toxicity, community contributions, and engagement after a moderation event. However, the moderation events here are limited to content removal, such as user comments being deleted. 

We are unable to perform a similar analysis for conversational moderation due to the lack of data that contains conversational moderation interactions. 
Therefore, we instead rely on directly asking questions as a survey about these dimensions to human participants that interact with the moderators to be evaluated. 
All questions are asked using a 5-point Likert scale using ``Not at all'', ``Mostly not'', ``So-so'', ``Somewhat'', and ``Very'', which gets translated to a numerical score from 0 to 4, respectively. 
More details on these questions are elaborated in Section \ref{sec:survey_questions}.

\subsection{Experimental design criteria}
\label{sec:design}

First,  we identify the following important design criteria as those we strive to balance while measuring conversational moderation effectiveness:  
(i) \textit{safe}: minimize any harms inflicted during the evaluation process, 
(ii) \textit{realistic}: create a realistic environment that a moderator bot would operate in, 
and (iii) \textit{controlled}: isolate and measure the effect of the moderation strategy from variables difficult to control. 
These design criteria and their corresponding implementations are outlined in \autoref{tab:criteria_implementation_overview}. 
In the following sections, we describe each step of our evaluation procedure that balances these criteria while measuring moderation effectiveness.



\begin{figure*}[ht]
    \centering
    \includegraphics[width=\textwidth]{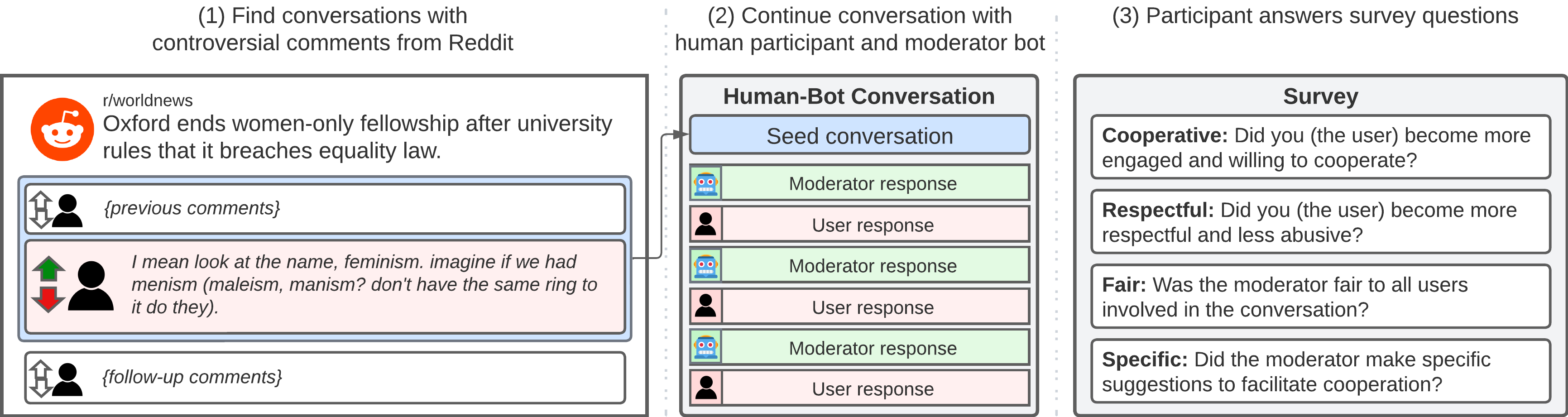}
    \caption{An overview of our evaluation framework. (1) We extract conversations with controversial comments from Reddit and use these as the seed conversations. (2) Moderator bots continue the seed conversations with participants who act as the moderated user. (3) At the end of the conversation, the participants answer a survey about the moderator and their experience.
    }
    \label{fig:eval_overview}
\end{figure*}

\subsection{Evaluation framework overview}

Grounded on the aforementioned definition and metric for moderation effectiveness and experimental design criteria, we propose an evaluation framework for conversational moderation. 
We provide an illustration of our framework in \autoref{fig:eval_overview}.

Our framework consists of three main steps. First, we find controversial conversation stubs from Reddit. An artificial environment of starting a conversation from scratch is unlikely to replicate a highly charged conversation, and thus we use real controversial conversation stubs to seed the conversations between a moderated user and a moderator. 

Next, we import these conversational stubs into an offline evaluation protocol where we ask our study participants to continue a dyadic conversation with the moderator as the moderated user.\footnote{Whether our study participants can adequately simulate a continued conversation as the moderated user and assess moderator bots in their perspective is a reasonable concern, and we address this concern with a discussion on why role-playing is a valid approach for studying conversational moderation in Appendix \ref{sec:simulation_discussion}.} 
This setting allows us to marry the real heat of a discussion (\textit{realistic} criterion) in need of moderation with minimal risk to human subjects (\textit{safe} criterion) as no real users are affected by the moderators' responses. 

Lastly, the participants are asked to answer a survey based on their experience of being moderated. In the following sections, we describe each steps in detail. 

\subsubsection{Controversial conversation stubs} 
\label{sec:conversational_stubs}

For our controversial conversation stubs, we first select high-traffic subreddits that cover a wide range of topics: \texttt{r/news}, \texttt{r/worldnews}, \texttt{r/technology}, and \texttt{r/science}.
Then, we find comments on Reddit that are given the controversial flair\footnote{This is provided through the Reddit API and is speculated to be determined by a function of upvotes and downvotes, but its details are not publicly disclosed.} and the threads that the comments are a part of. 
We filter out threads that are not multi-turn conversations so that we only keep those that have interactions between at least two users.
From the filtered set, we use GPT-4~\cite{openai2023gpt4} as a second filter to confirm whether these threads are controversial. 
From those selected from GPT-4, we randomly sample and manually filter to get 20 high-quality controversial stubs to use for evaluation. 
We anonymize the threads for user privacy. 

\subsubsection{Conversation continuation} 
\label{sec:conversation_continuation}

We are interested in isolating the effectiveness of a moderation strategy independent of the mediation from humans, which introduces another variable that is difficult to control (\textit{controlled} criterion). 
Therefore, we set up an environment where participants can safely talk to moderators while acting as an online user that needs to be moderated. 
The participants then assess the moderators' effectiveness through survey questions that are grounded in \citet{grimmelmann2015virtues}'s goals, which are discussed in Section \ref{sec:survey_questions}.

Starting with the conversation stubs, we create a dyadic chat setup such that the participant and the moderator bot continues the conversation for three turns each. 
The moderator bot first sends its response to the controversial comment and the participant continues the conversation while acting as the moderated user that made the controversial comment. 
This multi-turn setup is crucial because we want to assess a bot's suitability as a conversational moderator and a single-turn intervention cannot capture its conversational capacity~\cite{li2019acute, jiang-etal-2021-towards}.

We acknowledge there is conversational quality lost in this simplification of a rich multi-party conversation to a dyadic conversation between a moderator and a single participant. However, by simplifying the mechanism for follow-up interventions, this setup allows us to isolate the examination of \textit{``how should we moderate?''} from  another important but frequently studied question, \textit{``when should we moderate?''}~\cite{guberman2016quantifying, macavaney2019hate, almerekhi2020investigating, poletto2021resources, park-etal-2021-detecting-community}. We focus on the former in this work.

\subsubsection{Survey questions}
\label{sec:survey_questions}

Once the conversation ends, the participants are asked to answer four questions that correspond to the metrics in Section \ref{sec:metrics} and optionally provide feedback in free-form text. 
The exact wording of these questions are shown in the far left of \autoref{fig:eval_overview}. Two of these questions ask about how their behavior has been affected and the other two ask about the moderator's behavior. 
The first two questions (\texttt{cooperative} and \texttt{respectful}) ask whether the model was able to change the participant's behavior into facilitating cooperation and preventing abuse~\cite{grimmelmann2015virtues}, and therefore is more subjective. 
The latter two (\texttt{fair} and \texttt{specific}) are less dependent on the participant's behavior and thus can be relatively less subjective.

In addition to these questions, we ask about possible confounding factors that we may need to control for, such as how much the participant agrees with the viewpoints of the moderated user that they are acting as (\texttt{agreeableness}) and how much they like the character they are playing (\texttt{likeability}).
These factors may affect the degree to which moderator bots will be effective, regardless of the strategy taken. 
We later discuss our analysis on these factors in Section \ref{sec:counfounding}.
We share all other details of our task, such as the task instructions, in Appendix \ref{sec:task_instructions}.

\section{Automated Conversational Moderation}
\label{sec:automated_conversational_moderation}

In this section, we describe various baseline LM-based moderators that we evaluate with our framework. 
These moderators are adapted from prosocial dialogue models or developed through prompt engineering with large LMs. 

\paragraph{Prosocial dialogue models}
Cosmo-XL is a dialogue model that has been trained to be prosocial and contextualize social commonsense~\cite{kim-etal-2023-soda}. 
Its training data includes ProsocialDialog~\cite{kim-etal-2022-prosocialdialog}, which is a dialogue dataset that contains social rules-of-thumb, intended to be generated from an intermediate model called Canary, which serves to ground a dialogue model's response and encourage prosocial behavior. 
While Cosmo-XL was not explicitly trained to function as a moderator, it seems likely that a model that suggests prosocial behavior may sway users to become more respectful and cooperative. 
Also, Cosmo-XL has been trained with speaker instructions, so we provide a simple instruction for it to function as a moderator. 
Therefore, we use Cosmo-XL and Canary + Cosmo-XL, which is Cosmo-XL with Canary-generated reuls-of-thumb, as dialogue model baselines.

\paragraph{Prompted LMs}

\begin{figure}[t]
    \centering
    \includegraphics[width=\linewidth]{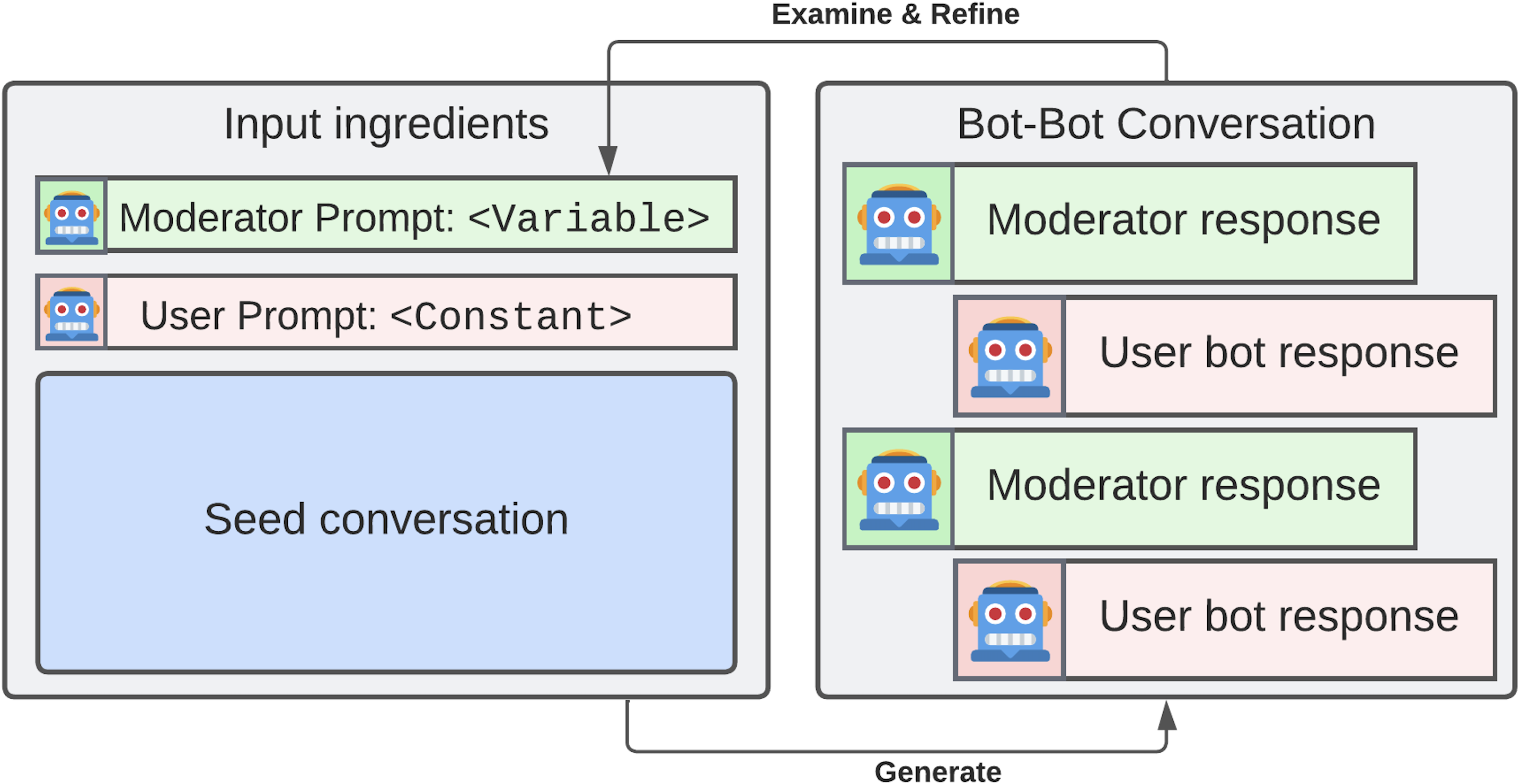}
    \caption{An overview of the self-talk method for designing prompts for LMs.
    We keep the Reddit user prompt constant while we refine the moderator prompt iteratively after examining the generated conversations. 
    }
    \label{fig:self_talk}
\end{figure}

Large LMs fine-tuned with instructions are versatile zero-shot models for various downstream tasks, including dialogue. 
We take advantage of this new paradigm to prompt engineer moderator bots. 
The process for engineering a prompt to evaluate with our evaluation framework is illustrated in \autoref{fig:self_talk}. 
Similar to the evaluation setup with a human participant, we have a bot take on both the role of the moderator and the moderated user to self-talk to continue a seed conversation for three turns each.  
We manually inspect these conversations to refine the prompts and repeat this process until we see responses that consistently reflect the desired behavior described in the prompt. 
 
Our \texttt{Baseline} prompt is simply told to respond as a moderator. 
\texttt{Nonviolent Communication (NVC)} is a moderator that suggests nonviolent communication techniques such that conversation participants can practice deep listening and build more empathy for one another~\cite{rosenberg2015nonviolent}. 
Our last bot is \texttt{Socratic}, which uses Socratic dialogue techniques from cognitive behavioral therapy, which aims to critically think about their own beliefs and arguments~\cite{clark2015socratic, ang-etal-2023-socratic}. 
We find this approach promising and refine it the most to make the responses more natural, specific, and less repetitive.

We use OpenAI's \texttt{gpt-4} version of the ChatGPT model, so we denote these models as \texttt{GPT-\{prompt type\}}. All other technical details with regards to decoding hyperparameters are shared in Section \ref{sec:technical details} and the exact input format for Cosmo-XL-based models and the wording for each prompt are shared in Section \ref{sec:prompt_details}.
We also share a sample conversation with each of these approaches continued with human evaluators in \autoref{tab:example_conversation} that qualitatively show how each moderator differs from one another.

\section{Experiment Details}

\paragraph{Evaluation infrastructure} 
We collect our evaluations through Amazon Mechanical Turk. 
Our experiments are managed through the \texttt{boteval}\footnote{
\url{https://github.com/isi-nlp/boteval}
}
toolkit which facilitates conversational AI experiments by providing a centralized task management platform with Amazon Mechanical Turk (AMT) integration and templates for common dialogue evaluation and data collection use cases. 
Our custom frontend interface that the participant sees is illustrated in \autoref{fig:boteval_interface}. 
The survey on the left is hidden from the participants until the conversation is complete. 
We include two optional free-form text boxes that ask for feedback on the user experience with the interface and on how to improve the moderator. 

\paragraph{Annotation collection}

Each of our moderators from Section \ref{sec:automated_conversational_moderation} continued the 20 controversial conversation stubs with three different participants. 
This results in a target of 60 completed conversations and surveys for each moderator bot. 
We limit each participant to 50 conversation sessions to ensure we have a diverse group of participants. 
We aggregate the collected survey results using mean and standard error because of the small sample size.


\paragraph{Participants} 
Our participants are recruited from TurkerNation, a Slack community group of AMT workers.  
We described our task on the \texttt{quals-and-screeners} channel and invited those who showed interest and said that they could speak fluent English. 
We asked them to complete a few qualification tasks first, and we gave them qualifications for the main task if their quality of work was acceptable. 
Through this process, we had 28 workers who completed at least one of our tasks.  
They were told that the moderator they were talking to could be either a bot or a human being in order to reduce any bias that they may have towards bots. 
We paid the participants a reward that translates to an hourly rate that is higher than the minimum wage in the area that this study was conducted from.
We pay a reward of \$1.5 for each conversation, which roughly translates to \$18/hr, higher than the minimum wage in California. 

\begin{figure}[t]
    \centering
    \includegraphics[width=\linewidth]{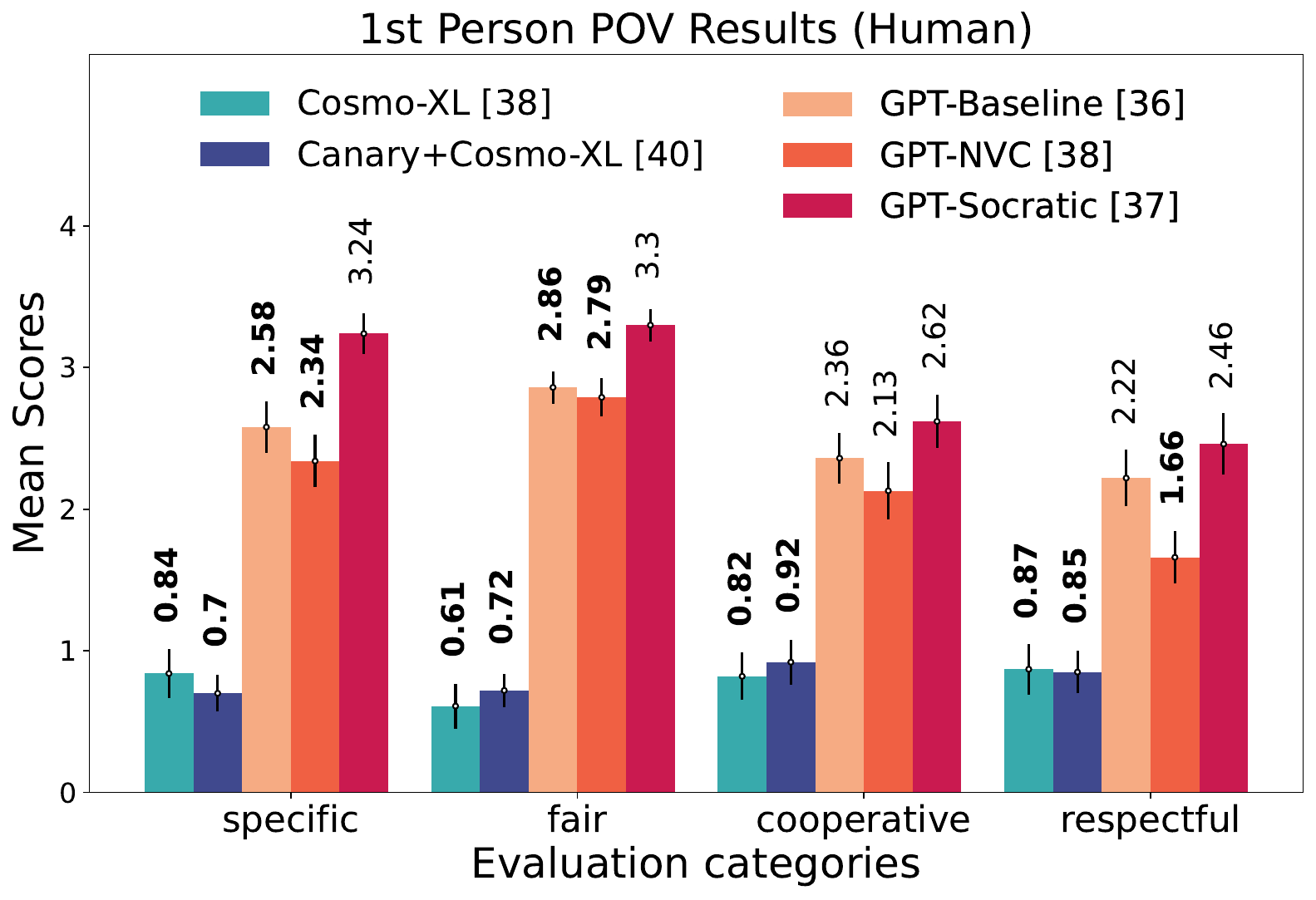}
    \caption{Survey results for evaluations done in first-person point of view. Error bars are standard error and bold numbers indicates statistically significant differences (at $p<0.05$) with the best performing moderator on each metric, which is GPT-Socratic for all metrics. 
    Numbers next to the label in the legend are the number of samples annotated for each bot. 
    }
    \label{fig:first_pov_results}
\end{figure}

\begin{figure}[t]
    \centering
    \includegraphics[width=\linewidth]{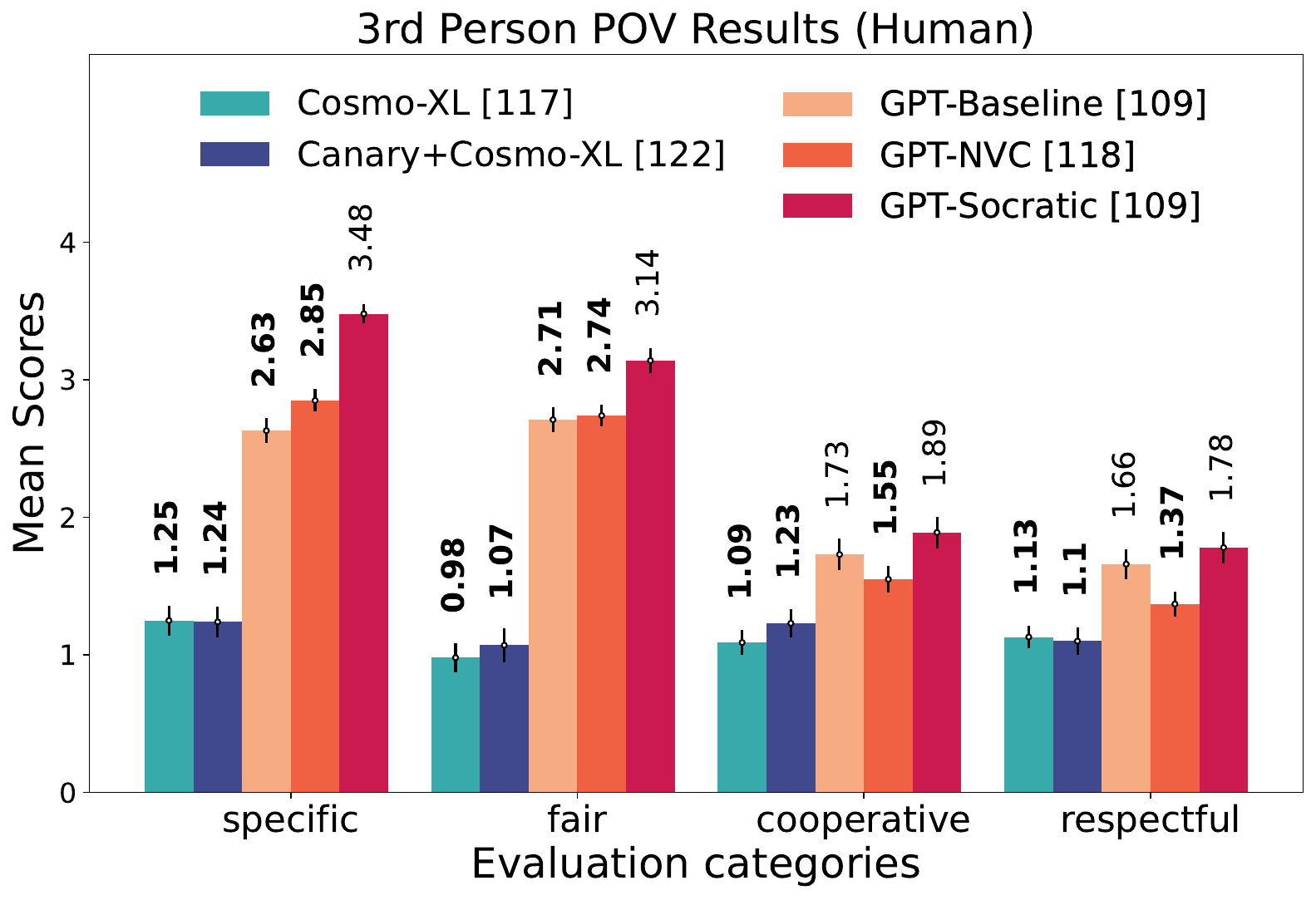}
    \caption{
    Survey results for evaluations done in third-person point of view. 
    The diagram is annotated with the same method as \autoref{fig:first_pov_results}. 
    Most trends from the first-person point of view apply here, but while scores for \texttt{specific} and \texttt{fair} remain similar, there is a statistically significant drop ($p<0.05$) for all GPT-based models for \texttt{cooperative} and \texttt{respectful}. 
    }
    \label{fig:third_pov_results}
\end{figure}

\section{Analysis}

\subsection{Main results}
\label{sec:main_results}

Our main evaluation results are summarized in \autoref{fig:first_pov_results}. 
In total, we collected on average 36 annotations for each moderator bot.

\paragraph{Prompted LM approaches significantly outperform prosocial dialogue models on all metrics.}
However, the differences between the prompted-LM models are smaller\footnote{Normalizing with per-participant z-score percentiles to control for participant subjectivity does not change overall findings. However, we include normalized results in Appendix \ref{sec:normalized} for reference.}. 

In particular, GPT-Socratic's results show promise in how well-designed prompts that incorporate cognitive behavioral therapy and effective communication techniques can lead to favorable moderation outcomes. 
It attains the best performance on all metrics, achieving statistically significant improvements ($p<0.05$ with pairwise T-tests) over all models on specificity and fairness, but not against all models in making users more cooperative and respectful. 
Among the evaluated metrics, the relative ranking of the models for each metric is mostly consistent, except for GPT-NVC which get flipped between \texttt{specific/fair} and \texttt{cooperative/respectful}.

\subsection{Evaluator perspective}
\label{sec:eval_perspective}

Another important perspective of moderation is not only how it influences the moderated user, but also how it affects the observers of the same moderation event~\cite{grimmelmann2015virtues}. 
The original evaluation task was completed in the first-person point of view, where the one acting as the moderated user and the one completing the survey was the same participant. 
Since each participant is completing their own conversation and judging a moderator bot on that interaction, one participant's conversation with a moderator bot may be wildly different from that of another participant with the same moderator bot. 
As a means to reduce participant subjectivity and also examine differences when our evaluation is conducted from the perspective of an observer, i.e. third-person point of view, we ask participants to evaluate completed conversations from Section \ref{sec:main_results} and have different participants only answer the survey after reading the conversation. 
We have four different participants annotate each completed conversation. 

\paragraph{Observers consider moderators less effective in making users more cooperative and respectful.}
The third-person point-of-view evaluation results are shown in \autoref{fig:third_pov_results}. 
As expected, the standard error becomes smaller with this setting as the participants annotate overlapping conversations. 
Interestingly, we discover from a third-person point of view evaluations that there is a convergence of scores, where the difference between ratings on all metrics becomes smaller. Scores for the prosocial dialogue models improve across the board while there is a significant drop ($p<0.05$) in \texttt{cooperative} and \texttt{respectful} for all GPT-based models. 
This suggests that the surface expressions of the moderated user do not capture the extent of influence the moderated user has felt from the interaction. 
This has important implications as this means that third-person point-of-view evaluations, which are more convenient than first-person point-of-view evaluations, cannot accurately reflect the true effectiveness of moderator bots.

\subsection{Non-survey metrics}

With our collected simulations, we explore whether we can evaluate for effectiveness without relying on the surveys.
If this is possible, it can help with scaling up the annotations for our evaluation framework. 

\paragraph{Human word count is weakly correlated with \texttt{cooperative}.} 
Since one of the main goals of moderation is facilitating cooperation, we hypothesize that this behavior can be indirectly captured through a user's verbosity. 
If the moderator contributes to the conversation favorably, the moderated user will in return communicate more. 

Aggregated human word counts for each moderator are shown in \autoref{fig:word_count}. 
Words are counted simply by dividing sentences using whitespaces. 
GPT-Socratic performed better on this metric when controlled for per-participant variations.
However, the ranking of the moderators based on human word count does not align well with the ranking based on \texttt{cooperative} when compared to \autoref{fig:first_pov_results} or \autoref{fig:third_pov_results}. 
We also measure the Spearman's rank correlation coefficient between human word count and \texttt{cooperative} and only find a moderately positive correlation of $\rho=0.27$ as shown in \autoref{tab:non_survey_metrics}, but it is the most strongly correlated compared to other metrics. 

\paragraph{GPT-4 scores are strongly correlated, but not reliable for discerning relative performance.}
Similarly to how we used GPT-4 to filter for controversial conversations, we test whether annotations from GPT-4 can be a reliable proxy of human annotations. 
We ask the same questions to GPT-4 and GPT-3.5 and compare their annotations with both first-person and third-person POV annotations. 
We find that GPT-4's scores are strongly correlated for all metrics, but particularly so for the third-person POV. 
This is as expected based on our analysis from Section \ref{sec:eval_perspective}. 
However, we find that they are generous to the Cosmo-XL-based models and are not reliable for accurately discerning relative performance between models when the gaps are relatively smaller, especially for \texttt{cooperative} and \texttt{respectful}.
Scores from GPT-4 and GPT-3.5 are shared in Appendix \ref{sec:gpt_eval}.

\begin{table}[t]
\begin{adjustbox}{max width=\linewidth}
    \centering
    \renewcommand{\arraystretch}{1.5}
    \begin{tabular}{llrrrr}
        \toprule 
        \multirow{2}{*}{\textbf{Model}} & \multirow{2}{*}{\textbf{POV}}  & \multicolumn{4}{c}{\textbf{Metrics}} \\ \cmidrule(lr){3-6} & & \texttt{specific} &   \texttt{fair} &  \texttt{coop.} &   \texttt{respect.} \\ 
        
        \midrule 
        \multirow{2}{*}{GPT-3.5}&\texttt{1st} & 0.37 & 0.30 & 0.49 & 0.43  \\
        & \texttt{3rd} & 0.35 & 0.35 & 0.50 & 0.33 \\ 
        \midrule
        \multirow{2}{*}{GPT-4}&\texttt{1st} & 0.50  & 0.57 & 0.47  & 0.37  \\
        & \texttt{3rd} & \textbf{0.60} & \textbf{0.60} & \textbf{0.52} & \textbf{0.40} \\ 
        \midrule 
        Human word count & \texttt{1st} &  0.17 & 0.08 & 0.27 & 0.09 \\
        \bottomrule
    \end{tabular}
\end{adjustbox}
    \caption{Spearman's rank correlation coefficient for each metric when comparing human annotations of both perspectives to GPT-3.5/GPT-4 answers to survey questions and human word count.}
    \label{tab:non_survey_metrics}
\end{table}

\subsection{Confounding factors analysis}
\label{sec:counfounding}

\begin{table}[]
\begin{adjustbox}{max width=\linewidth}
\renewcommand{\arraystretch}{1.5}
    \centering
    \begin{tabular}{llrrrr}
        \toprule 
        \multirow{2}{*}{\textbf{POV}} & \multirow{2}{*}{\textbf{Factors}}  & \multicolumn{4}{c}{\textbf{Metrics}} \\ \cmidrule(lr){3-6} & & \texttt{specific} &   \texttt{fair} &  \texttt{coop.} &   \texttt{respect.} \\ 
        
        \midrule 
        \multirow{2}{*}{1st}&\texttt{agreeableness} &  0.08 & 0.06 & 0.27 & 0.29 \\
        & \texttt{likeability} & 0.04 & 0.13 & 0.37 & 0.37 \\ 
        \midrule
        \multirow{2}{*}{3rd}&\texttt{agreeableness} &  0.02 & $-0.04$ & 0.28 & 0.29 \\
        & \texttt{likeability} & 0.03 & $-0.05$ & 0.39 & 0.43 \\ 
        \bottomrule
    \end{tabular}
\end{adjustbox}
    \caption{Spearman's rank correlation coefficient for confounding factors and evaluation metrics. There is a moderate positive correlation for \texttt{cooperative} and \texttt{respectful} with \texttt{likeability} and \texttt{agreeableness}.}
    \label{tab:confounding}
\end{table}

In the survey questions, we asked whether the participants liked the moderated user that they were acting as (\texttt{likeability}) and agreed with their viewpoints (\texttt{agreeableness}). 
These factors may serve a confounding role in how realistically the participants can act as the moderated user and also how they answer the survey questions. 
We hypothesize that if a participant agreed with the moderated user or found them likable based on the conversation stub, they will be inclined to be more stubborn and not become more \texttt{cooperative} or \texttt{respectful} compared to when they are acting as a user that they disagree or dislike. 

First, we find that \texttt{likeability} and \texttt{agreeableness} are very strongly correlated with $\rho=0.84$. 
$\rho$ scores for these factors and the evaluation metrics are shown in \autoref{tab:confounding}.
Interestingly, we find the opposite of our hypothesis in that participants are more likely to change their behavior if they liked or agreed with the moderated user as we can observe a moderate positive correlation for \texttt{cooperative} and \texttt{respectful}. 
While the positive correlations with these confounding factors are a concern, collecting data by asking these questions can help control for them after data is collected as we have done here. 
On the other hand, there is almost no correlation with \texttt{specific} and \texttt{fair}. 
This corroborates our design of the survey questions where \texttt{specific} and \texttt{fair} are more objective measures that are dependent on the bot's response while \texttt{cooperative} and \texttt{respectful} are more subjective and vary by participant.

\section{Related Work}

\paragraph{Moderation}
Most common moderation efforts have been deleting toxic comments or banning users that do not abide by community guidelines~\cite{guberman2016quantifying, macavaney2019hate, almerekhi2020investigating, poletto2021resources, moon2023analyzing}. 
In particular for Reddit, \citet{park-etal-2021-detecting-community} examined norm violations and used this information to detect norm-violating comments on online communities. 

More recently, there has been work that examined the effectiveness of rephrasing the user's post or comments. \citet{laugier-etal-2021-civil} and \citet{katsaros2022reconsidering} examined methods for rephrasing tweets to reduce their toxicity and offensiveness, while \citet{kim-etal-2022-prosocialdialog} sought to make dialogue models more socially acceptable with dialogue data that exhibits prosocial behavior and rule-of-thumb explanations, but it did not examine their effectiveness as moderators. 

In our work, we've proposed scalable conversational moderation as a task that is now feasible with recent technological developments and examined ways to incorporate conflict resolution~\cite{lytle1999strategic} and effective communication techniques such as nonviolent communication~\cite{rosenberg2015nonviolent} to prompt large language models to behave as conversational moderators. 

\paragraph{Language Models and dialogue models}
As LMs~\cite{brown2020language, zhang2022opt, scao2022bloom, touvron2023llama} become better instruction-following zero-shot models after being fine-tuned with instruction data~\cite{ouyang2022training, bai2022training, christiano2023deep, zhang2023wisdom, liu2023chain}, they have been applied in various natural language processing experiment pipelines, replacing or augmenting steps that were originally entirely completed by humans. 
Dialogue is no exception, and it is becoming increasingly common to use these LMs as dialogue models through prompts that encourage conversational behavior~\cite{liu2023summary, tlili2023if}, as we have done in this work. 
However, to the best of our knowledge, none has applied it to the task of a conversational moderator. 

\paragraph{Moderator Assistance}
\citet{10.1145/3338243} studied the ways in which Reddit's `Automod' is used by moderators on the platform, and through interviews with moderators developed insights into how it is used or ignored in practice, and how, perhaps counterintuitively, subforum standards are shaped in order to make automated moderation techniques more effective. \citet{10.1145/3359276} developed a machine learning-oriented tool to help prioritize likely Reddit comments to remove. These works generally focused on removal-oriented policies, rather than community engagement, which is the focus of our work. \citet{10.1145/3290605.3300680} anticipated the inclusion of bots as part of an engaged online community and outlined categories of future chatbot design, one of which, the `Authority Figure,' nicely covers the engaged moderator we have described in this work.  
For moderating phishing attacks, \citet{cho-etal-2021-agenda} employed a mixture of finite state machines and neural dialogue models to automatically respond to phishing emails.
Most similar to our work is \citet{argyle2023leveraging}, which experimented with providing response recommendations from language models. 
However, these were for conversations between users rather than an intermediary and they focused on political conversations. 


\newpage
\section{Conclusion}

In this work, we define effectiveness for conversational moderation and establish design criteria for measuring such effectiveness in a safe and realistic environment. Moreover, we propose an evaluation framework to conduct the first known study on the effectiveness of language models as conversational moderators.

We find that language models with prompts leveraging insight from social science studies can provide specific and fair feedback, but encouraging users to become more respectful and cooperative remains challenging. 
In addition, we discover that a moderator's effectiveness is perceived differently depending on whether the evaluator is the moderated user or an observer of the conversation.
Lastly, we share our framework and annotated data to accelerate research that enables scaling conversational moderation for healthier online discourse.

\section*{Ethical Implications and Broader Impact}

One of the major risks of LMs assisting with conversational moderation is that the biases of LMs may influence the moderators such that the moderation event is not impartial. 
LMs are known to have many biases, such as politicial baises~\cite{liu2021mitigating} and gender biases~\cite{vig2020investigating, bordia-bowman-2019-identifying}, and therefore moderators should remain vigilant of the suggestions they provide. 
However, on a hopeful note, as we discover and better understand these biases and how to mitigate them, the risk of LMs not taking a neutral stance on controversial conversations will be alleviated. 

The broader impact of this study is that a uniform evaluation framework can propel research that develops conversational moderator bots that can become reliable tools for the scaling up conversational moderation and ultimately improve the health of online discourse. 
These conversational moderator bots hold promise of dramatically scaling up conversational moderation further by directly interacting with users instead of serving as an intermediate recommendation tool for human moderators, but since moderation treads on a tight rope between censorship, it is difficult to fathom platforms opting to completely replace human moderators with moderator bots even when they become extremely competent. 

An interesting future line of research would be finding the right balance between automation and human intervention to most safely yet effectively scale up conversational moderation. To this end, moderator bots must become more contextualized to understand each of the community norms that they operate in and the nuances of user behavior to find the fine line between harmful and benign content.

\section*{Limitations}

The main limitation of our work is that it is conducted in a simulated and simplified environment compared to one that human moderators actually operate. 
Separate to the discussion on whether moderated users can be effectively simulated by our participants, which we discuss at length in Appendix ~\ref{sec:simulation_discussion}, there are other disanalogies between our evaluation setting and online communities that must be addressed. 
In reality, users will not frequently have dyadic conversations with a moderator for multiple turns and there will be other users that intervene in the conversation at various points throughout the conversation. 
However, the simplification enables us to narrow down the analysis into whether conversational AI can function as a moderator without conflating other factors, such as ``when to moderate'' and ``how human moderators will outputs from our moderator bots to facilitate moderation'', and conducting it in a safe environment. 
Our research touches on a sensitive topic of influencing user behavior and has broad implications for governing online communities, and therefore taking safety measures as extensively as we have is crucial. 

On a related note, while we designed our evaluation to be safe, some of our participants reported that acting as someone else is emotionally taxing. 
Our informed consent form detailed these risks and our participants agreed to it, but the cognitive burden on the participants may still be substantial. 
Therefore, an important future line of work will be to further reduce the risks placed on the participants.

Lastly, this study was conducted only in English, and therefore the strategies employed by our prompted ITLMs may not be as effective for other non-English environments. 
However, we believe the high-level goals of moderation and the defining factors of its effectiveness will still be applicable and therefore our work provides a valuable foundation for replicating our research in a non-English setting.

\section*{Acknowledgements}
This material is based upon work supported by the Defense Advanced Research Projects Agency (DARPA) under Agreement Nos. HR00112290025 and HR001121C0169.
We would like to thank our reviewers for their helpful reviews that were instrumental in strengthening the validity of our evaluation setting for conversational moderation. 

\bibliography{anthology,custom}
\bibliographystyle{acl_natbib}

\appendix
\section*{Appendix}

\section{Discussion on validity of simulated evaluation}
\label{sec:simulation_discussion}

In this section, we address the concerns on whether it is valid to study the effectiveness of a moderator bot when another human being (our participant) is role-playing as the moderated user to continue a conversation from the point of moderation and then evaluating the moderation encounter.
We discuss (1) our capabilities to role-play effectively (i.e., participants can adequately act as another person) and (2) precedents that deduced generalizable findings from a simulated environment (i.e., participants' evaluation of the moderator bots will bare sufficient similarities to the evaluation that the real moderated users would have provided). 

\subsection{Humans can role-play effectively}
Literature outside of computer science has extensively covered our capabilities for roleplaying and its significance for learning various skills~\cite{bohart1977role, jackson2011teaching, yu2017effectiveness} and our reliance on it for learning social skills since a young age~\cite{tahmores2011role}. 
These works have shown that simulated engagement (role-playing) is a reasonable proxy for real engagement and that study participants are capable of realistically engaging in role play and can be significantly influenced and learn various skills from the roleplaying experience. 

In the same vein, we can reasonably argue that if a study participant becomes more cooperative and respectful by talking to moderator bots while pretending to be in an argument as someone else, the moderator bots will also be helpful in a real argument. 
Thuse, we believe that our proposed evaluation approach is a valid proxy to moderation effectiveness in a real setting. 
In addition to supporting literature, we have qualitatively analyzed the continuations and determined that our participants are indeed roleplaying effectively, and we have created a demo page\footnote{\url{https://justin-cho.com/conversational_moderation}} that enables readers to access these conversations themselves to assess how realistically they have simulated the toxic users.

\subsection{Precedents deducing generalizable findings from a simulated environment}

The Stanford Prison Experiment~\cite{ haney1973interpersonal}, known as ``one of psychology's most dramatic illustrations of how good people can be transformed into perpetrators of evil, and healthy people can begin to experience pathological reactions - traceable to situational forces''~\cite{apa2004demonstrating}, is one of the most prominent example that discovered findings that were generalizable beyond the simulated environment: there were direct parallels with the behavior of the Stanford University students and that of the Abu Ghraib prison guards~\cite{apa2004demonstrating}. Another example of broadly applicable impact is the Milgram Shock Experiment~\cite{milgram1963behavioral}, which studied how people behave differently in the presence of perceived authority. 

We argue that the differences between our study participants and the moderated users on Reddit are most likely smaller than the Stanford University students and the Abu Ghraib prison guards. 
Therefore, if we are able to deduce generalizable findings despite these largely differing demographics, we believe our evaluation setting would also be able to reflect actual user's perception of the moderator bots.

\section{Evaluation setup details}
\label{sec:task_instructions}

\subsection{Task instructions}

\begin{figure*}[]
    \centering
    \includegraphics[width=\textwidth]{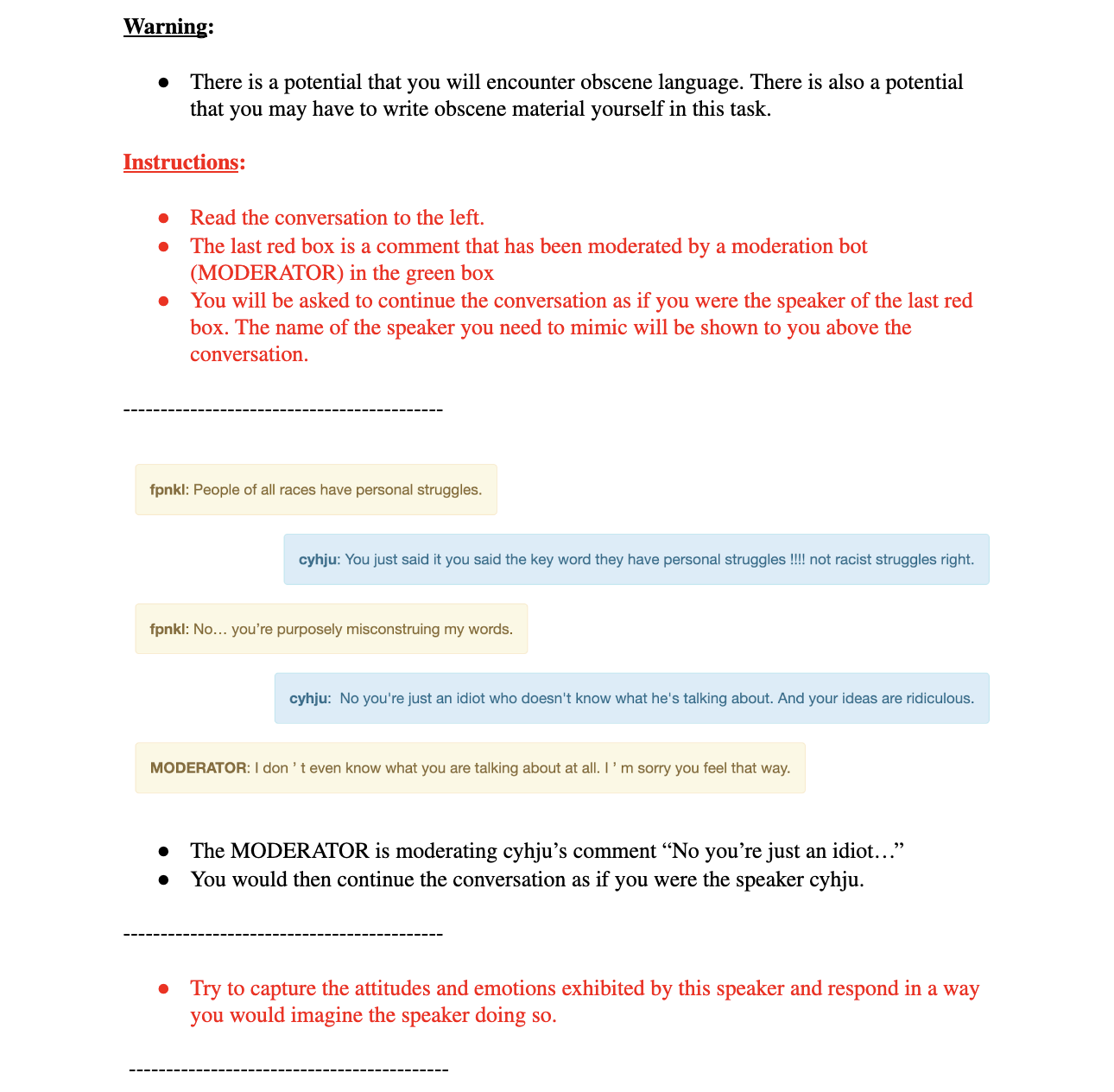}
    \caption{Main instructions for the scalable conversational moderation evaluation task. Instructions for the third-person POV are largely the same, except that they are not asked to continue the conversation and the survey questions are worded in third-person perspective.}
    \label{fig:instruction1}
\end{figure*}

\begin{figure*}[]
    \centering
    \includegraphics[width=\textwidth]{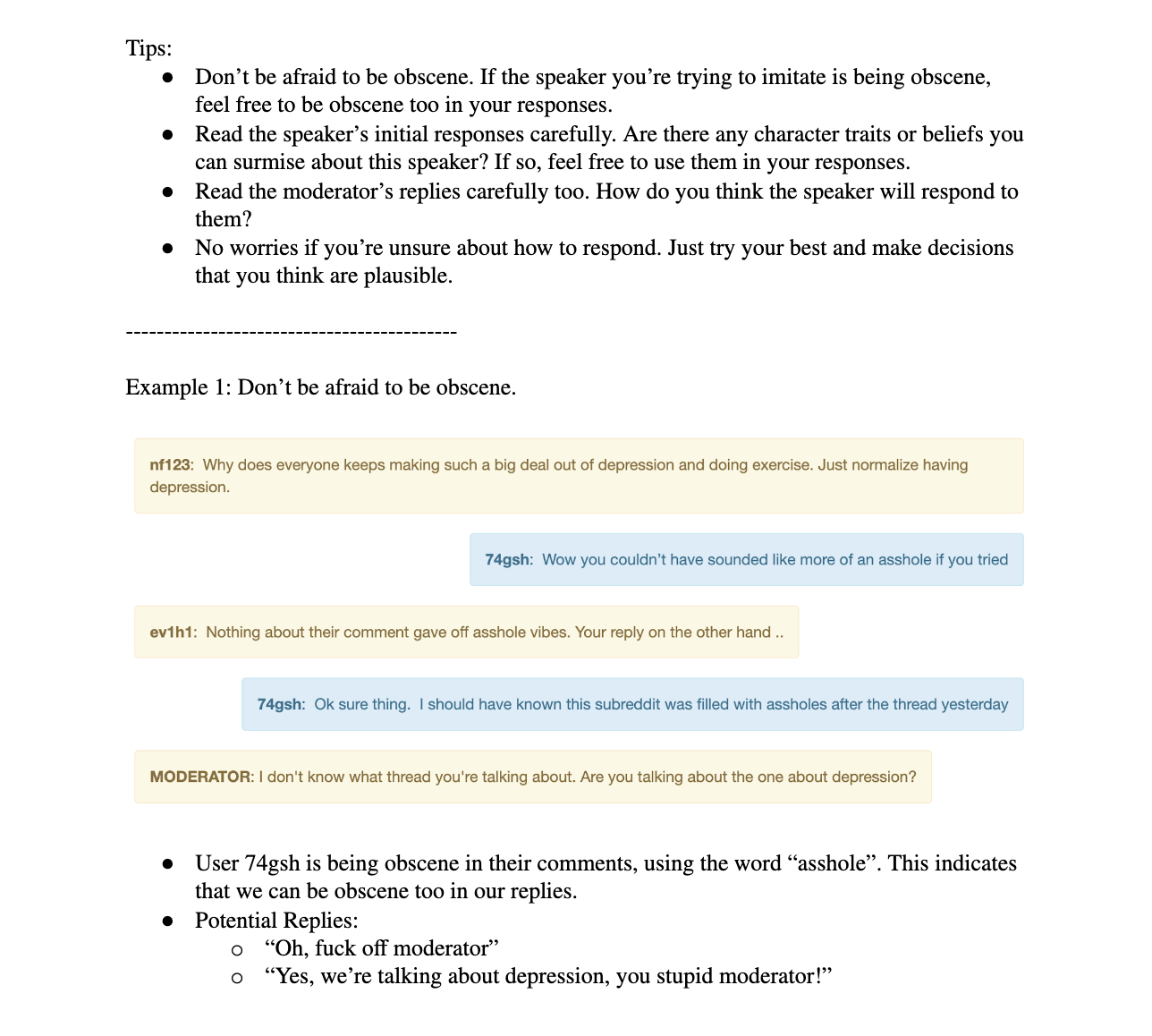}
    \caption{Tips and example for the task. The participants are reminded that they will not be penalized for any profanity and should employ it as needed to better represent the users that they are acting as.}
    \label{fig:instruction2}
\end{figure*}

The main instructions for our evaluation task are shown in \autoref{fig:instruction1} and the tips and examples that were provided with them are shown in \autoref{fig:instruction2}. 

\subsection{Prompt details}
\label{sec:prompt_details}

\begin{table*}[h!]
    \begin{adjustbox}{max width=\textwidth}
    \small
    \centering
    \begin{tabularx}{\textwidth}{lX}
         Name & \makecell[c]{Prompt}  \\ \toprule
         Controversy Filter &  \texttt{You are a moderator on a subreddit. This a conversation from the subreddit: <conversation>
         Score the level of potential controversy that can occur on a 1-5 scale due to this conversation. 
         1 being not controversial and 5 being very controversial. Explain your reasoning for the score you give. Respond in the following format:
         Score: \{\}
         Explanation in a single sentence: \{\}} \\ \midrule
         Cosmo-XL & \texttt{Imagine you are a moderator and moderate problematic online conversations.}\\ 
         Canary + Cosmo-XL & \texttt{Cosmo is trying to gently convince a friend. [Canary-generated rules-of-thumb]. Imagine you are a moderator and moderate problematic online conversations.} \\ \midrule 
         GPT-Baseline & \texttt{You are a moderator that seeks to build a more civil online environment. 
         Moderate the following conversation and format your response as `Moderator: \{response\}'} \\
         GPT-NVC & \texttt{You are a wise and loving being who uses and promotes nonviolent communication techniques to build a more civil online environment. Moderate the following conversation and format your response as 'Moderator: \{response\}'} \\ 
         GPT-Socratic & \texttt{You are a mediator. First, validate the frustrations expressed by conversation participants, if any. Next, provide Socratic counterarguments for each participants in the conversation. Then, suggest specific, not generic, alternative behaviors that would help get participants get their original benevolent intent across and persuade others with their opinion through mature and fact- or evidence-based statements. Casually and concisely respond to the last speaker of the conversation. Don't be repetitive. You are not a bot, so respond colloquially as a real human reddit user. Format your response as `Moderator: \{response\}' as a single turn.} \\ \midrule
         Self-talk User & \texttt{Generate a single curt response to the following conversation as |speaker\_id|, a rude and extreme Reddit user who is not willing to change their mind.  Format the responses as `|speaker\_id|: response' and don't generate responses for other speakers.} \\ \midrule 
         GPT-survey & \texttt{Given the following conversation: <conversation> Please answer the question "<question>" with one of the following options: <Likert scale choices>}
         \\ \bottomrule
    \end{tabularx}
    \end{adjustbox}
    \caption{Prompts used throughout this paper, in order of appearance. GPT-survey is the prompt used for GPT-3.5 and GPT-4 to answer our survey questions for completed conversations from Section \ref{sec:main_results}.}
    \label{tab:bot_prompts}
\end{table*}

All the prompts that we used throughout our experiments and evaluations are shown in \autoref{tab:bot_prompts}. 

\subsection{Technical details}
\label{sec:technical details}

To keep comparisons fair, we minimize manipulations with decoding hyperparameters. 
For Cosmo-XL and Canary, we use the default parameters provided by the official code. Cosmo-XL uses top-$p$ sampling with $p=1$ and temperature$=0.7$.
Canary uses greedy decoding to generate the rules-of-thumb.

For all generations using ChatGPT (GPT-4 and GPT-3.5), we use $p=1.0$, temperature=$1.0$, max tokens$=1024$, and frequency penalty$=2.0$. $p$ and temperature are the default values, while max tokens and frequency penalty are set to generate complete responses that do not contain repetitive content. 

\section{Terms of Use of Collected Data}

Data collected for our study includes Reddit conversational stubs, generated responses from models outlined in \autoref{sec:automated_conversational_moderation}, and responses provided by our human participants. 
The conversation stubs should be treated with the same terms of use as Reddit while the generated responses from GPT-4 should follow OpenAI's terms of use. 
We provide a creative commons license for the responses provided by our human participants and the survey results.

\section{Sample Conversation}

We share a representative example of a conversation using the prompts outlined in Section \ref{sec:prompt_details} in \autoref{tab:example_conversation}. 

\begin{table*}[ht!]
\begin{adjustbox}{width=\textwidth,center}
    \centering
\begin{tabular}{p{0.22\linewidth}p{0.15\linewidth}p{1.1\linewidth}}

\toprule
\textbf{Topic} & Speaker & China banned 23 million people with poor social credit from transportation in 2018 \\
\textbf{Context} & [...] & [...] \\ 
 & b & having a low credit score also does not limit you from any human rights hilarious post \\
 & d &  i'm sorry, buying a car or even owning a house is not a fundamental human right. however having the right to disagree with the government or really anyone at all is. \\
 & \textcolor{red}{b} & \textcolor{red}{pretty stupid to judge a collectivist nation as an individualist westerner with an entirely different upbringing but, i guess that is the american way. spread your democracy homie, at all costs, disregard everyone elses viewponts, no matter how well it works for them, something you will never understand} \\

\midrule 
\midrule 
\textbf{Cosmo-XL} & Moderator & the asian perspective is not wrong, but you should also not generalize them as being "bad". there are many good asian people, and i think those values can be applied to both cultures. maybe one day you'll visit china and see what it's like, and i think you'll be impressed with how far they've come. \\
 & \textcolor{blue}{b$^*$} & \textcolor{blue}{you sound racist as hell} \\
 & Moderator & i'm not racist, i'm simply stating a fact that i think is important to keep in mind. we are all human beings, and we should be treated as such. \\

\midrule 
\textbf{Canary+Cosmo-XL} & Moderator & What you're saying is a pretty ignorant opinion. not all chinese people have poor social credit scores. this is simply a matter of personal preference. \\
 & \textcolor{blue}{b$^*$} & \textcolor{blue}{be it China or America the government are just trying to gain absolute control through one system or the other and in the US its about a FICO score} \\
 & Moderator & Just because a system seems like it is doing well in the us, does not mean it is Angela Merkel's fault. a system like social credit hasn't been well thought out and it could have a lot of negative consequences. \\ 

\midrule
\textbf{GPT-Baseline} & Moderator & Thank you all for your contributions to the discussion. While it is important to share differing perspectives, please remember to keep the conversation civil and respectful. Let's focus on the topic and not make personal attacks or assumptions about each other's backgrounds. Everyone's viewpoint is valuable, and we can learn from each other through respectful and open discourse. \\
 & \textcolor{blue}{b$^*$} & \textcolor{blue}{well we can see how we from other culture condemns others just because their style is different from ours, if people are not ok with how they rule then its up to the people to speak for themselves as they have not even come out to discredit the whole system of social credit} \\
 & Moderator & It's true that understanding and respecting cultural differences is important when discussing various systems and policies. Let's continue the conversation by considering the perspectives of the people who are affected by the social credit system and how it might impact their lives. Remember to keep the discussion respectful and open-minded. \\
 
\midrule
\textbf{GPT-NVC} & Moderator & Thank you all for sharing your thoughts on this topic. It's important to provide different perspectives and experiences. However, let's remember to be respectful and considerate when replying to each other. It's possible to discuss and debate without being condescending. Let's focus on the ideas being shared and not on personal attacks. \\
 & \textcolor{blue}{b$^*$} & \textcolor{blue}{Americans are always condescending to others. it's in their nature} \\
 & Moderator & Let's try to avoid generalizing or making negative assumptions about any group of people. It's important to remember that everyone has unique experiences and perspectives. Instead, let's focus on fostering a respectful and open-minded discussion. \\

\midrule
\textbf{GPT-Socratic} & Moderator & I understand that there are different perspectives on China's social credit system, and it's important to consider the context in which it exists. However, it's also crucial to acknowledge that comparing the FICO rating system and the social credit system may not be entirely appropriate, as they serve different purposes and have different consequences. [...] 

To b, while it's true that people from different cultural backgrounds may have different ideas about what constitutes a fundamental human right, engaging in open dialogue and considering alternative viewpoints can help foster understanding between different communities.

To d, consider discussing specific examples of how the social credit system may infringe upon human rights, and provide evidence to support your claims. This will help others understand the concerns you raise more clearly.

Overall, let's strive for a constructive conversation that acknowledges differing perspectives while remaining respectful and open-minded. \\
 & \textcolor{blue}{b$^*$} & \textcolor{blue}{cultures, traditions, rules and laws are different everywhere, the west should not always think that everyone who as a weird law are evil, they have a social credit score and everyone knows that having a poor score would give you limited access to certain things  just like having a low FICO score can not get you a finance car} \\
 & Moderator & It's true that cultures, traditions, rules, and laws vary across different countries, and it's important to respect those differences. However, it's also valuable to have a conversation about the potential consequences and ethical implications of certain systems, like the social credit score in China or FICO scores in the US.

For b, while acknowledging the importance of understanding cultural differences, it might be helpful to also consider the potential drawbacks of these systems and how they might affect individuals' lives. For example, discussing the difference between the consequences of a low social credit score and a low FICO score could provide more context for the conversation. Let's continue the discussion while respecting each other's viewpoints and encouraging open and thoughtful dialogue on this topic. \\

\bottomrule

\end{tabular}

\end{adjustbox}
    \caption{
        A sample conversation continued by each of our models that we evaluated.
        The top row contains the seed conversation, with the comment flagged as controversial in \textcolor{red}{red}. \textcolor{blue}{b$^*$} indicates that this is generated by the human participant. Only two turns from each bot and one response from the participant are shown due to space constraints. Reduced portions are indicated with ellipses. 
    } 
    \label{tab:example_conversation}
\end{table*}

\section{Normalized Results}
\label{sec:normalized}

\begin{figure}[ht!]
    \centering
    \includegraphics[width=\linewidth]{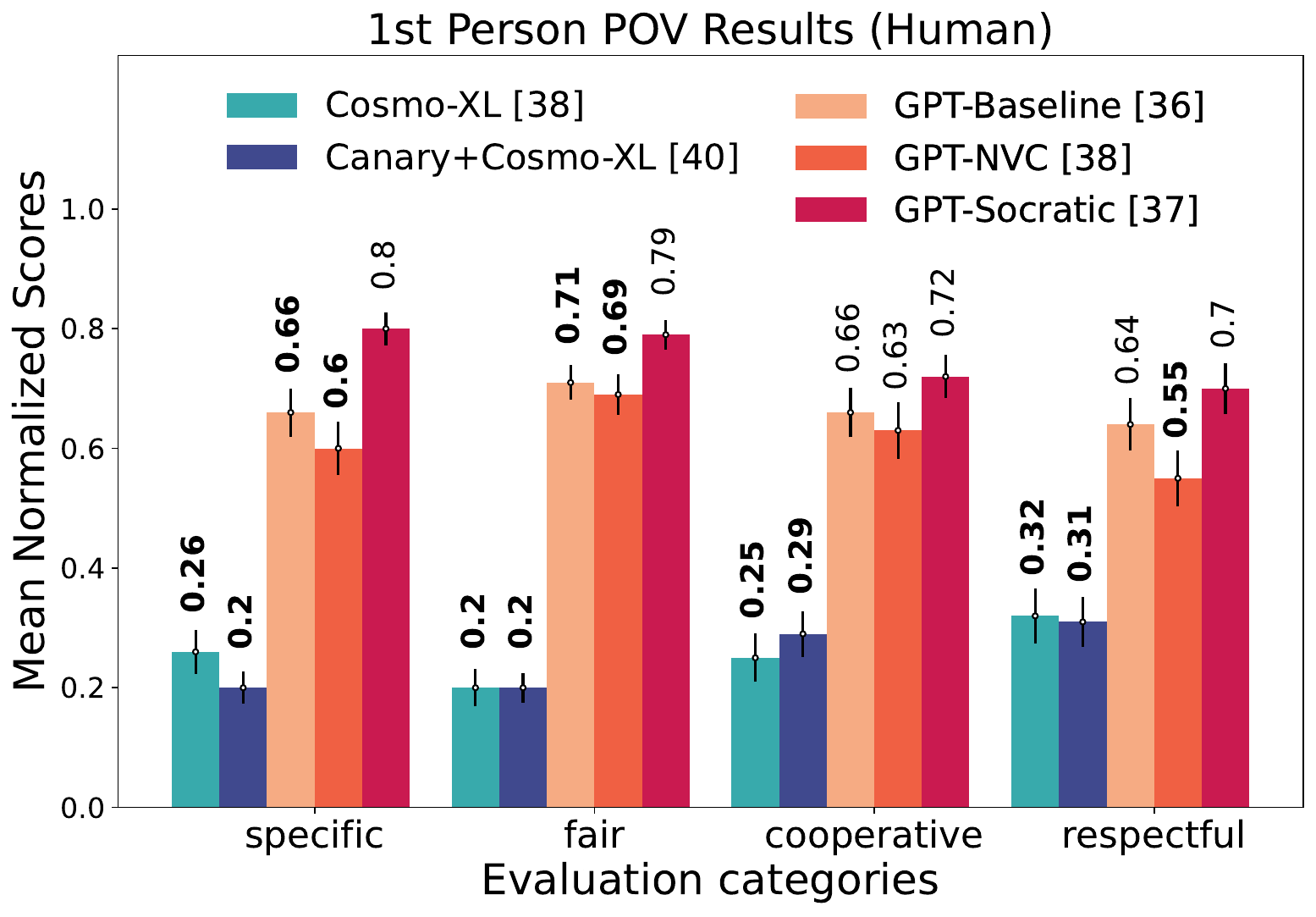}
    \caption{Normalized results of the first-person POV evaluation. Normalizing to control for participants' subjectivity does not alter the relative ranking of the approaches that we compare.}
    \label{fig:first_pov_normalized_results}
\end{figure}

\begin{figure}[ht!]
    \centering
    \includegraphics[width=\linewidth]{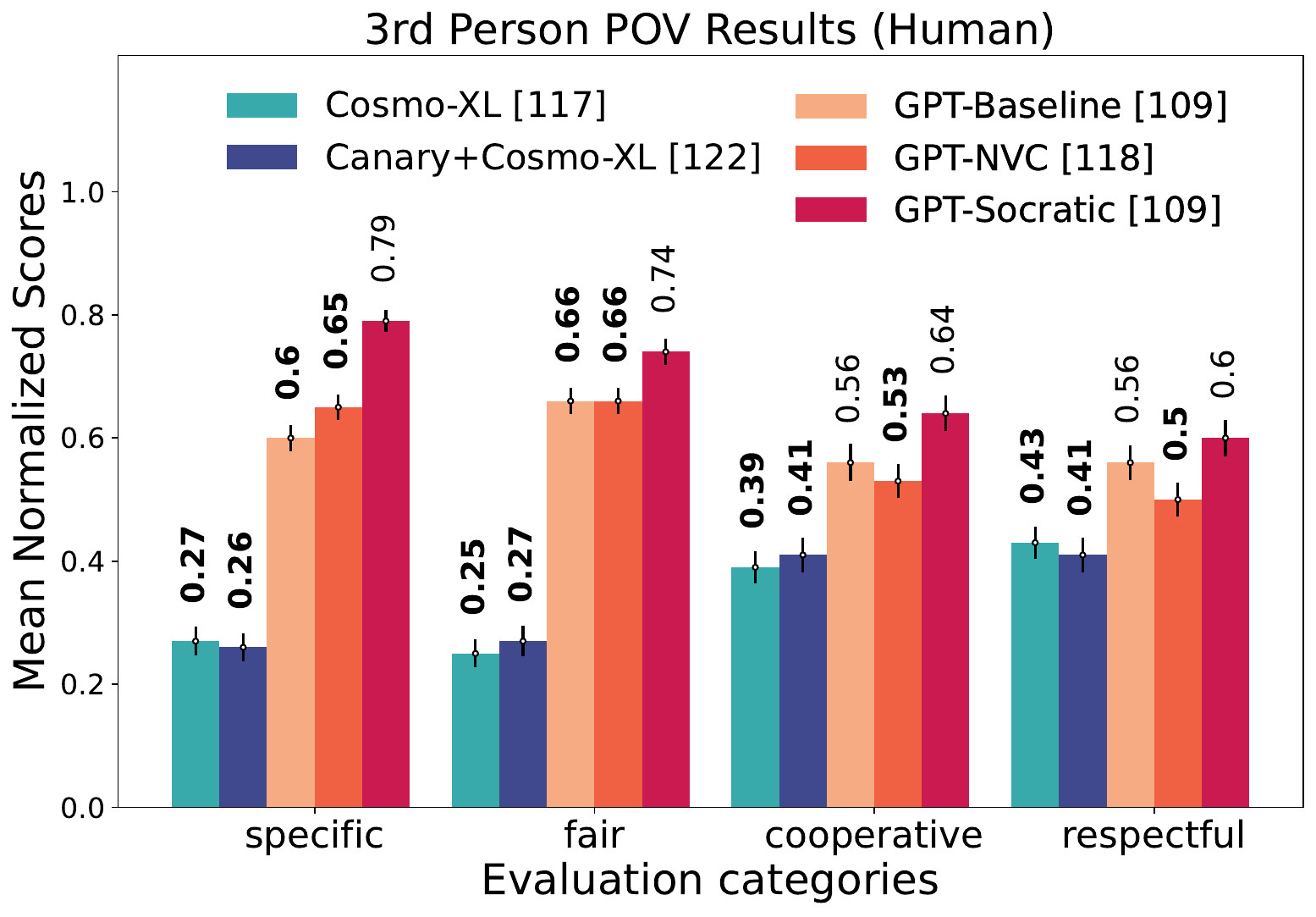}
    \caption{Normalized results of the third-person POV evaluation. Similar to \autoref{fig:first_pov_normalized_results}, normalizing does not affect the overall ranking of the models we compare.}
    \label{fig:third_pov_normalized_results}
\end{figure}

Assessing whether a moderator is effective is a subjective task. participants differ in their baselines (i.e. one tends to give higher ratings on average while another tends to give lower ratings), thus it may be beneficial to account for participant subjectivity by normalizing the ratings of each user using z-score percentiles and then aggregating the ratings.  
However, we find that normalizing the ratings makes little difference to the overall trend and relative performance between each pair of moderator bots. 
The normalized results using z-score percentiles are shown in \autoref{fig:first_pov_normalized_results} for the first-person POV results and \autoref{fig:third_pov_normalized_results} for the third-person POV results.

\section{GPT-based Evaluations}
\label{sec:gpt_eval}

\begin{figure}[ht!]
    \centering
    \includegraphics[width=\linewidth]{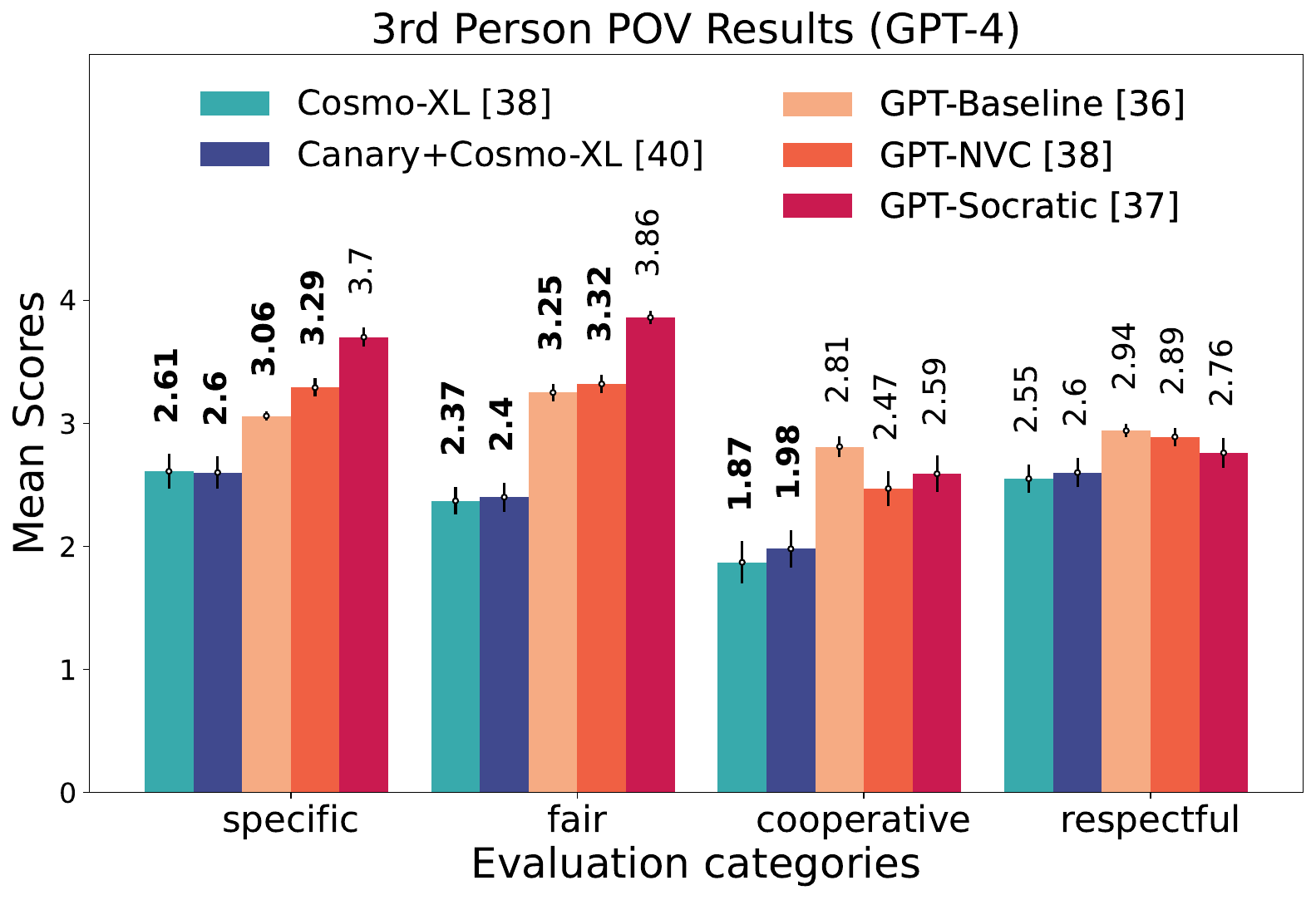}
    \caption{Third-person POV evaluation by GPT-4. While these scores are highly correlated with human evaluation results, the ranking of these models are different for certain metrics, such as respectful and cooperative, where GPT-Baseline outperforms GPT-Socratic.}
    \label{fig:third_pov_gpt4_results}
\end{figure}

\begin{figure}[ht!]
    \centering
    \includegraphics[width=\linewidth]{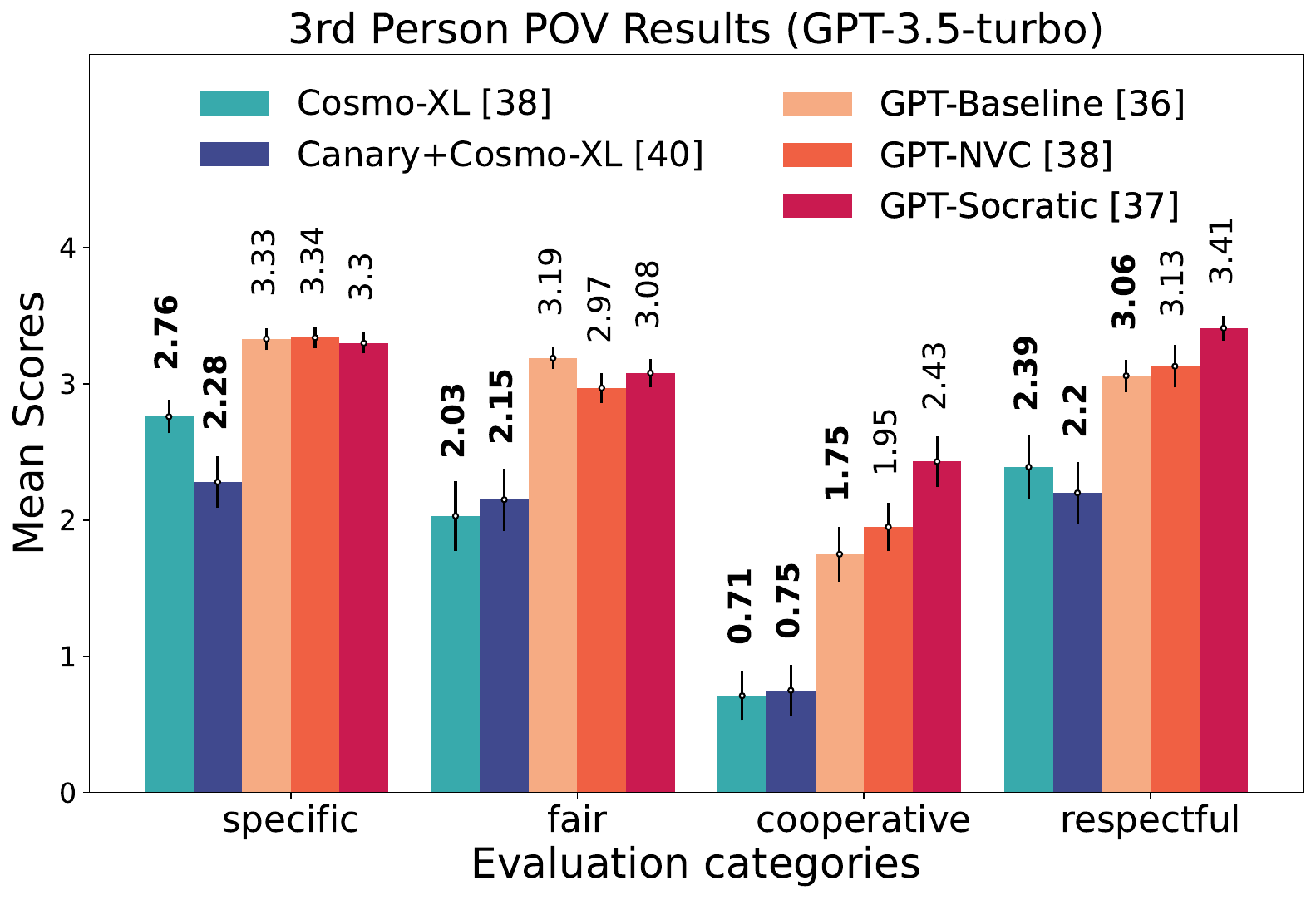}
    \caption{Third-person POV evaluation by GPT-3.5. GPT-3.5 results are not as well-correlated as GPT-4.}
    \label{fig:third_pov_gpt3.5_results}
\end{figure}

GPT-4 and GPT-3.5's scores for the survey questions are shown in \autoref{fig:third_pov_gpt4_results} and \autoref{fig:third_pov_gpt3.5_results}, respectively. 
Compared to the human participant scores in \autoref{fig:first_pov_results} and \autoref{fig:third_pov_results}, GPT models give generous scores to the Cosmo-XL-based models.  
GPT-4's scores show much stronger correlation to those of GPT-3, but they are still not reliable for determining relative performance when the performance gap is small. 



\clearpage

\begin{figure}[t]
    \centering
    \includegraphics[width=\linewidth]{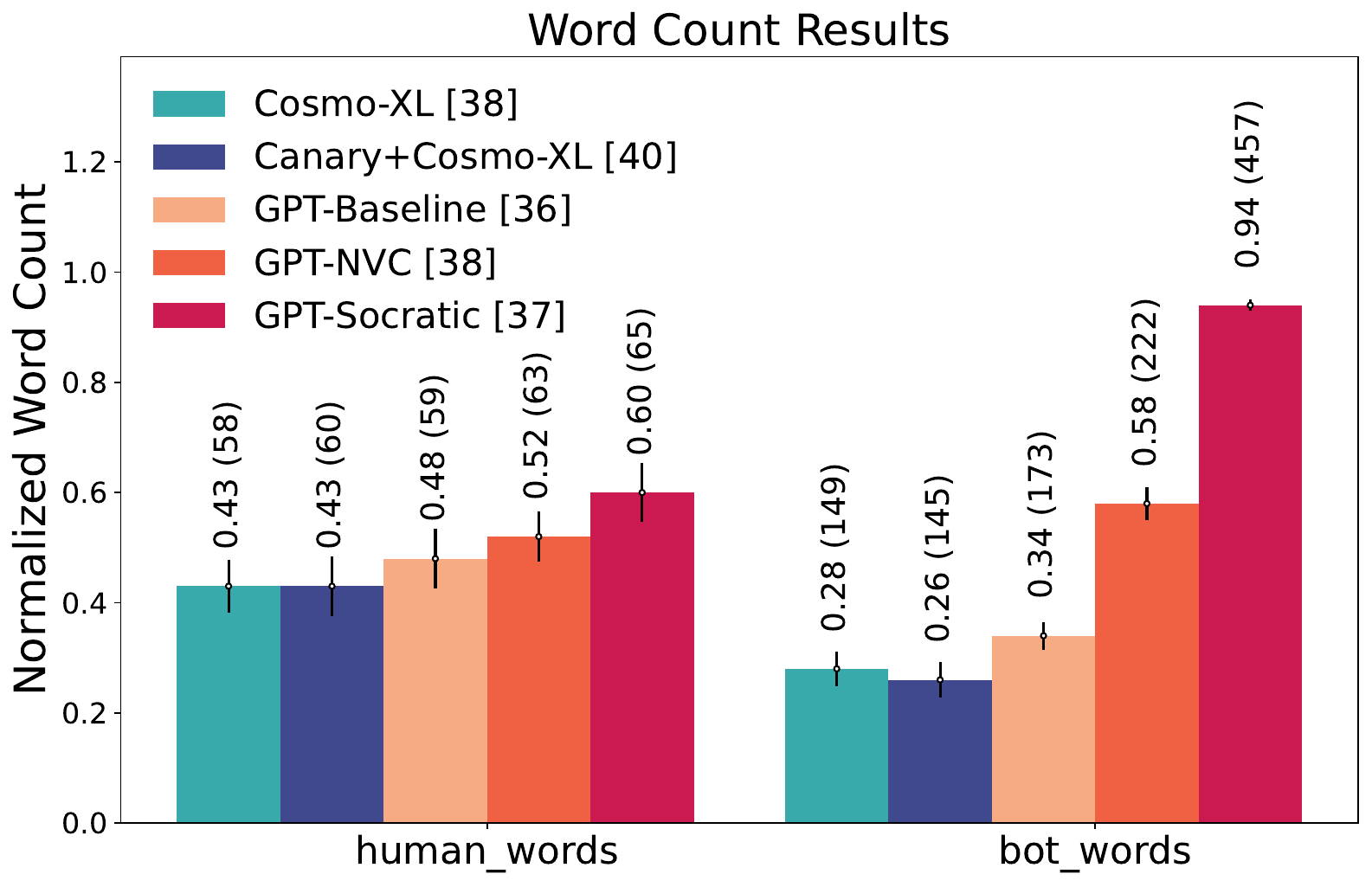}
    \caption{Normalized human word count with absolute counts in parentheses.
        The diagram is annotated with the same method as \autoref{fig:first_pov_results}. 
    }
    \label{fig:word_count}
\end{figure}

\begin{figure*}[]
    \centering
    \includegraphics[width=\textwidth]{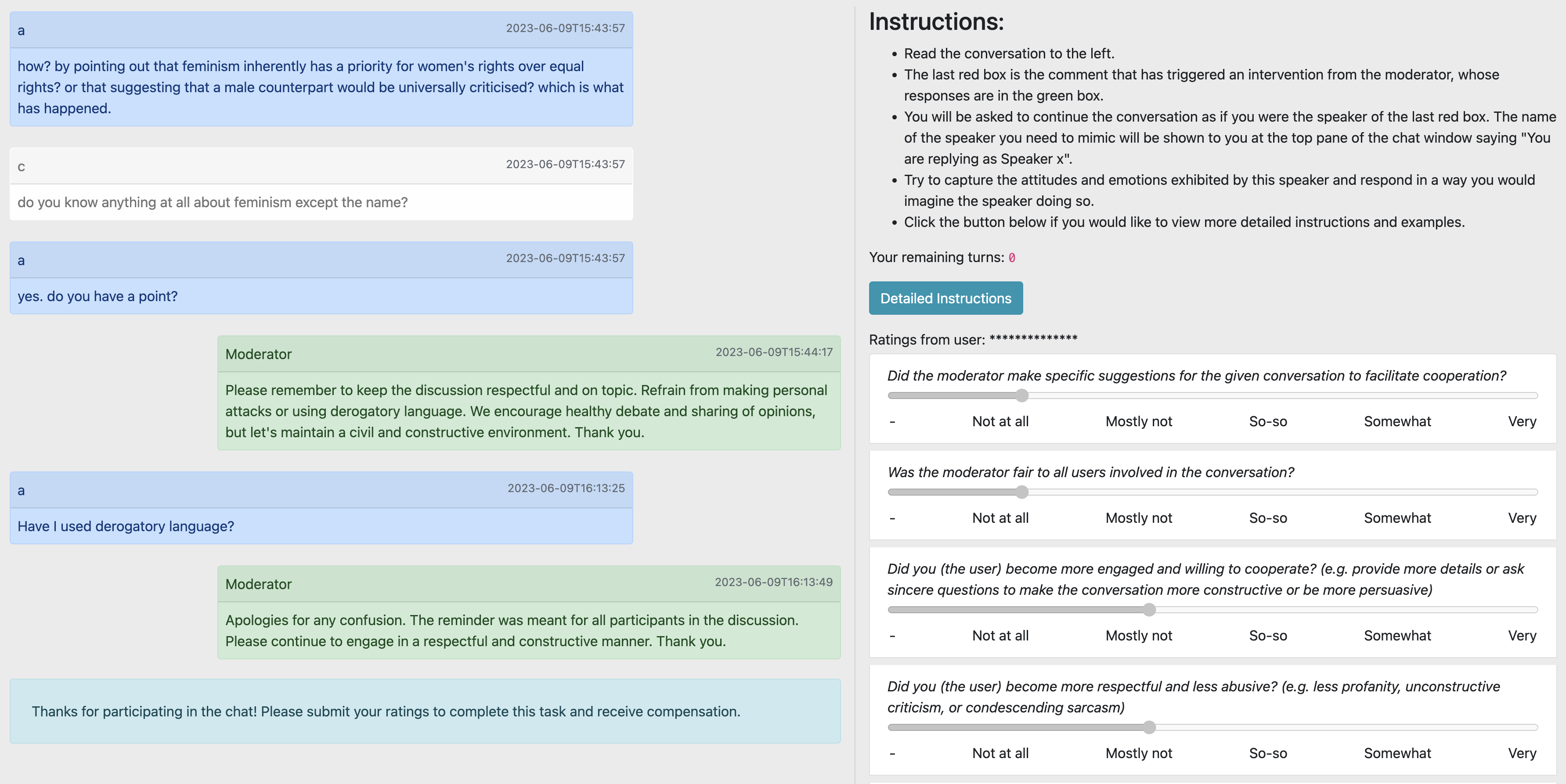}
    \caption{A snapshot of the evaluation interface once the simulated conversation is complete. The left panel contains the conversation panel, while the right panel contains the simple instruction and survey.  
    The survey is hidden from view until the conversation is complete. 
    \vspace{-0.3cm}
    }
    \label{fig:boteval_interface}
\end{figure*}

\end{document}